# A robust deep learning-based damage identification approach for SHM considering missing data

Fan Deng [1,2,3], Xiaoming Tao[1,2,3], Pengxiang Wei[1,2,3], Shiyin Wei[1,2,3,*]

1 Key Lab of Smart Prevention and Mitigation of Civil Engineering Disasters of the Ministry of Industry and Information, Technology, Harbin Institute of Technology, Harbin 150090, China.
2 Key Lab of Structures Dynamic Behavior and Control of the Ministry of Education, Harbin Institute of Technology, Harbin, 150090, China.
3 School of Civil Engineering, Harbin Institute of Technology, 150090, Harbin, China
Corresponding to: shiyin.wei@hit.edu.cn

## Abstract

Data-driven method for Structural Health Monitoring (SHM), that mine the hidden structural performance from the correlations among monitored time series data, has received widely concerns recently. However, missing data significantly impacts the conduction of this method. Missing data is a frequently encountered issue in time series data in SHM and many other real-world applications, that harms to the standardized data mining and downstream tasks, such as condition assessment. Imputation approaches based on spatiotemporal relations among monitoring data are developed to handle this issue, however, no additional information is added during imputation. This paper thus develops a robust method for damage identification that considers the missing data occasions, based on long-short term memory (LSTM) model and dropout mechanism in the autoencoder (AE) framework. Inputs channels are randomly dropped to simulate the missing data in training, and reconstruction errors are used as the loss function and the damage indicator. Quasi-static response (cable tension) of a cable-stayed bridge released in 1st IPC-SHM is employed to verify this proposed method, and results show that the missing data imputation and damage identification can be implemented together in a unified way.

## 1. Introduction

Big data of the in-service bridges are collected by Structural Health Monitoring (SHM) systems, in which lies structural performance that can be mined with data mining methods. Due to the promising of mining inherit patterns in data, machine learning and pattern recognition have been widely employed in the data mining and forms the data-driven method in SHM. In these data-driven approaches, structural performance is always exhibited in the correlation pattern in data [1]. As the currently most popular tool in artificial intelligence and data mining, artificial neural networks, especially its deep version (deep neural networks), provide an effective and efficient way to approximate the nonlinear mapping for the correlation mining. Recent years have seen an emerged development of applying artificial neural networks to conduct damage identification and condition assessment in the SHM community[2,3].

However, the proposed methods seldomly consider the occasion of missing data, which is a frequently encountered problem in SHM and other real-world applications [4–10] due to the sensor fault and made a well-trained model highly un-robust. Generally, missing data in SHM can be divided into 3 types: discrete missing at random time points, continuous missing of continuous time points, and continuous missing of the whole channel. Missing data introduce incomplete and nonstandard data format, thus affecting the data-driven methods for SHM.

Missing data imputation is thus developed to estimate the missing values from the available data, and to impute the missing data to form the standard inputs of the data-driven model[11]. Model-based methods and the deep learning-based methods are divided. Model-based methods develop a model (usually a regression model, or a statistical model) to consider the relationships among the datasets and reconstruct the missing data with the estimated statistical model, such as compressed sensing (CS) based, singular value decomposition (SVD) based, likelihood based, and K-nearest neighbor based, etc. Chen et. al. develops a distribution-to-warping function regression model of the the distributions of various channels based on the functional transformation technique[12]. Deep learning-based methods develops a block-box model of the spatiotemporal correlationship among multi channels of the monitoring time series and reconstruct the missing value to minimize the reconstruction error. Various DNN models a discussed such as recurrent neural networks (RNN), denoising autoencoders, convolutional neural networks (CNN), and generative adversarial neural networks (GAN), etc. Tang et. al. develops Group sparsity-aware CNN model for the continuous missing data imputation, where CNN is employed to generates the base matrix and the optimization in the group-sparsity reconstruction[13].

While most research focus on the missing data of missing type (a) and type (b), missing type (c) continuous missing is rarely considered. Moreover, most imputation methods are introduced as the data pre-processing and normalization procedure of the downstream tasks, and statistical/spatiotemporal correlations among the observed data is used in these methods[3]. However, these methods do not add any additional information during imputation, thus do not help the downstream tasks. For damage identification task in SHM, we need to infer whether damaged or not only by the information in the observed data.

Therefore, in this paper, we proposes a robust deep learning-based damage identification approach for SHM considering missing data. Autoencoder framework is employed to learn the inner relationship among monitoring data of different channels and to extract hidden representations, and LSTM is employed to construct the encoder and decoder module. The data reconstruction error is employed as the loss function and the damage indicator. In the training process, channels of the inputs are randomly dropped to simulate the missing data, and the LSTM-structured autoencoder model tries to reconstruct the data of all channels. And the SHM for the cable tension data, which was released in the 1st IPC-SHM (International Project Competition for Structural Health Monitoring) [14], is used as validation.

This paper is organized as follows. Section 2 first introduces the basic modules, i.e., the dropout mechanism, the LSTM cell, and the autoencoder framework, thereafter, proposes the model used in this paper that combines the basic modules. Section 3 then introduces the opensource dataset, the preprocessing procedure, and the implementation details of the proposed method. Section 4 discusses the results, including the proposed method, and the conventional DNN model.

## 2. Methodology

### 2.1 DNN and Dropout mechanism

A conventional deep neural network (DNN)[15] consists of three parts: the input layer, the output layer, and the hidden layers, as illustrated in Fig. 1 (a). While the input layer and output layer both has only one layer of units, hidden layers can have numbers of layers, and the number of the hidden layers is the 'depth' of the DNN model. Each layer contains numbers of units, and the number of units in each layer is the 'width' of the DNN model. Units of different layers are densely connected, and information flows forward from the lower layers to the upper layers, from the input layers to the output layers, therefore, this type of connection is also named as the feedforward neural networks. This process writes as:

$$h_i^l = \sigma\left(b_i^l + \sum_j w_{ij}^l h_j^{l-1}\right) = \sigma\left(\boldsymbol{b}^l + \boldsymbol{W}_i^l \boldsymbol{h}^{l-1}\right) \quad (1)$$

Where $\boldsymbol{h}^l = [h_1^l, \cdots, h_m^l]$ is the hidden state of the $l$th hidden layer, $h_i^l$ is the hidden state of the $i$th unit, and $\boldsymbol{h}^0 = \boldsymbol{x}$ is the input dataset; $w_{ij}^l$ is the weight of the $i$th unit in $l$-th layer and the the $j$th unit in $l$-1th layer, and $b_i^l$ is the bias; and $\sigma(\cdot)$ is the nonlinear activation function.

The information flows from the dataset $\boldsymbol{x}$ in the input layer to the predicted $\hat{\boldsymbol{y}}$ in the output layer for a DNN with $L$ hidden layers writes:

$$\hat{\boldsymbol{y}} = f(\boldsymbol{x};\boldsymbol{\theta}) = f^{L+1}(\boldsymbol{x}) = o\left(f^L\left(\cdots f^2\left(f^1(\boldsymbol{x})\right)\right)\right) \quad (2)$$

Where $f^l(\cdot)$ represents the nonlinear function of the $l$th hidden layer, $o(\cdot)$ is the output function, $\boldsymbol{\theta} = \{\boldsymbol{W}^1, \boldsymbol{b}^1, \cdots, \boldsymbol{W}^{L+1}, \boldsymbol{b}^{L+1}\}$ is the parameter to be learned in DNN. Given the target $\boldsymbol{y}$, a loss function that evaluates the prediction performance of the DNN model can be defined, e.g., the Mean Square Error (MSE) illustrated in Eq. (3).

$$L(\hat{y}_n, y_n) = L(f(x_n;\boldsymbol{\theta}), y_n) = \frac{1}{2}\sum_{n=1}^{N}\|f(x_n;\boldsymbol{\theta}) - y_n\|^2 \quad (3)$$

$$\boldsymbol{\theta} \leftarrow \boldsymbol{\theta} - \alpha \nabla L_{\boldsymbol{\theta}}$$

$L$ is a function of parameters $\boldsymbol{\theta}$ and $N$ is the total number in the dataset. A DNN model learns to predict the target $\boldsymbol{y}$ by adjusting its parameters $\boldsymbol{\theta}$ to minimize the loss $L(\hat{\boldsymbol{y}}, \boldsymbol{y};\boldsymbol{\theta})$. Gradient descent method is usually adopted for the parameter updating, the gradient $\nabla_{\boldsymbol{\theta}} L$ w.r.t. parameters $\boldsymbol{\theta}$ is calculated and backpropagated to parameters of all layers base on Eq. (2-3) and the chain rule. This training process is known as the Backpropagation (BP) algorithm.

Compared to shallow neural networks with hand-crafted features[16], deep learning is designed to learn more effective representations that extract the nonlinear relationships hidden in data through end-to-end training. However, parameter space can be extremely huge for a deep neural network with dense connections, inducing the training hard and overfitting issue. Dropout mechanism is thus proposed and is now a standard technique for training deep neural networks[17].

As illustrated in Fig. 1(b), dropout randomly ignores some units during training by enforcing the weights as 0 to reduce the connects in the neural network with a probability of $p$ [18], which can be expressed as:

$$h' = \begin{cases} 0 & \text{with probability } p \\ \dfrac{h}{1-p} & \text{otherwise} \end{cases} \quad (4)$$

Where $h$ and $h'$ represent the original the dropout hidden state respectively. It's obviously that the expectation $E[h'] = E[h]$, therefore, dropout is equivalent to add unbiased noises on the hidden units. Dropout is usually employed as a regularization to avoid overfitting and can be viewed as a type of ensemble learning[18]. Recent research proves that dropout in early and late stage of the training process is useful to avoid both overfitting and underfitting[19].

In the missing data imputation tasks, the dataset may have several random missing channels, thus making some units in the input layers zero. The randomly missing channels in the dataset will harm a well-trained network. Here we introduce the dropout mechanism in the input layers in the training process by randomly ignoring the units in the input layer to simulate the missing data in the inputs, we enforce the model to learn the invariant patterns in a missing data condition.

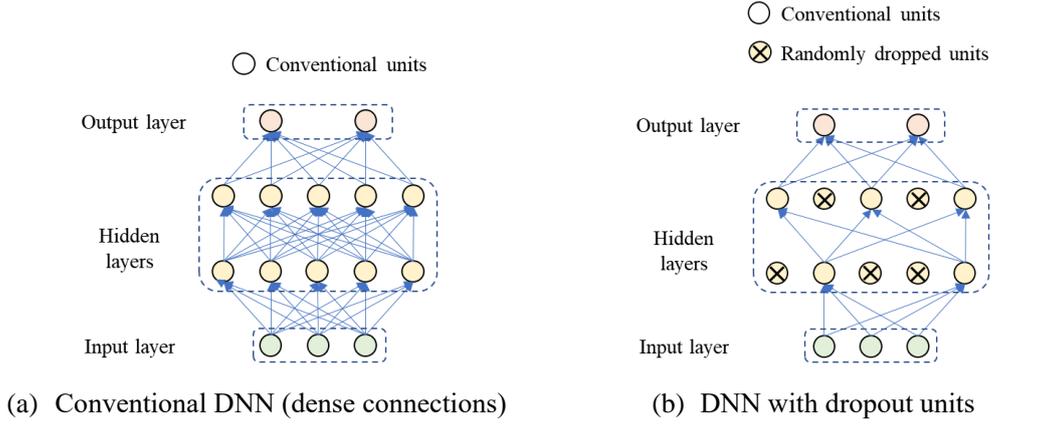

(a) Conventional DNN (dense connections)   (b) DNN with dropout units

Figure 1. Conventional DNN and dropout (arrow denotes the information flow direction)

## 2.2 LSTM

In DL tasks, the input and the output are usually specified according to the real applications, most DL models differ in the architecture of hidden layers. Temporal relation in time series data is usually hard to be modelled by the conventional DNN; recurrent connection that map outputs of earlier step to the later step is introduced to the hidden layers to model the temporal correlation. DNN with recurrent connections is the Recurrent Neural Network (RNN), as illustrated in Fig. 2 (a), where the red arrow represents the recurrent connections. Deep RNN model with multiple layers is usually illustrated in the folded expression in Fig. 2(b). where the rectangular blocks represent the hidden layers, and the cyclic arrows are the recurrent connections.

Long-short term memory (LSTM) is proven to be an efficient framework to model sequence data thus been employed as the correlation model in this study[20–22]. In LSTM model, the hidden layers are LSTM cells with gate units as illustrated in Fig. 4 (c). The updating of LSTM unit and its parameters are illustrated in Eq. (6) – (7).

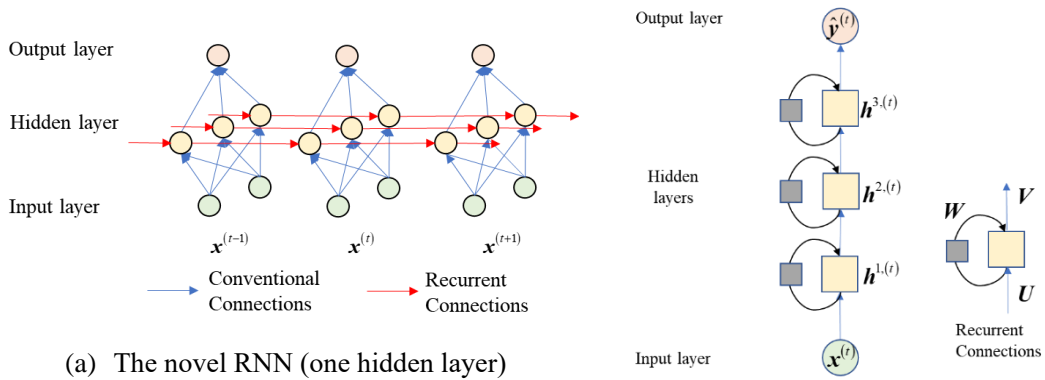

(a) The novel RNN (one hidden layer)

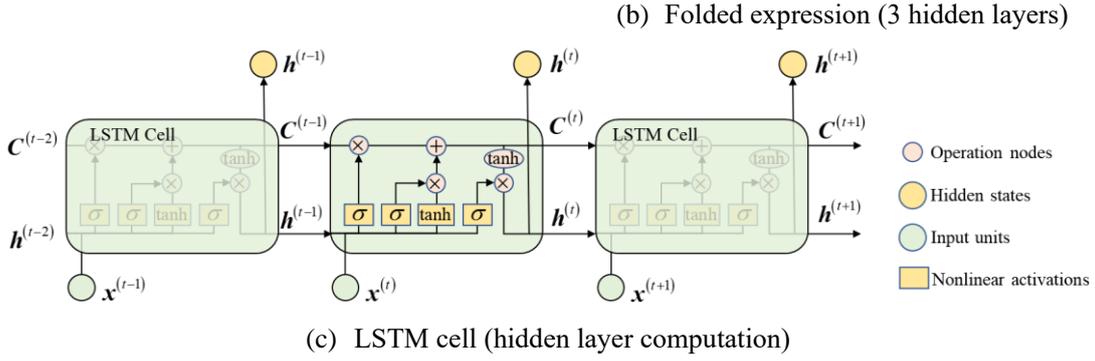

(b) Folded expression (3 hidden layers)

(c) LSTM cell (hidden layer computation)

Figure 2. RNN and LSTM architecture

$$f^{(t)} = \sigma\left(\mathbf{W}_f \mathbf{h}^{(t-1)} + \mathbf{U}_f \mathbf{x}^{(t)} + \mathbf{b}_f\right)$$
$$i^{(t)} = \sigma\left(\mathbf{W}_i \mathbf{h}^{(t-1)} + \mathbf{U}_i \mathbf{x}^{(t)} + \mathbf{b}_i\right) \quad (5)$$
$$o^{(t)} = \sigma\left(\mathbf{W}_o \mathbf{h}^{(t-1)} + \mathbf{U}_o \mathbf{x}^{(t)} + \mathbf{b}_o\right)$$

States updating:

$$\tilde{C}^{(t)} = \tanh\left(\mathbf{W}_C \mathbf{h}^{(t-1)} + \mathbf{U}_C \mathbf{x}^{(t)} + \mathbf{b}_C\right) \quad (6)$$

Outputs:

$$C^{(t)} = f^{(t)} * C^{(t-1)} + i^{(t)} * \tilde{C}^{(t)}$$
$$\mathbf{h}^{(t)} = o^{(t)} * \tanh\left(C^{(t)}\right) \quad (7)$$

Where $\mathbf{W}_f, \mathbf{W}_i, \mathbf{W}_o, \mathbf{W}_c, \mathbf{U}_f, \mathbf{U}_i, \mathbf{U}_o, \mathbf{U}_c, \mathbf{b}_f, \mathbf{b}_i, \mathbf{b}_o, \mathbf{b}_c$ are the recurrent weights, input weights and bias, $*$ represents for the elementwise product of matrices, and $\sigma(\cdot)$ is the Sigmoid activation function.

## 2.3 AE: AutoEncoder

Deep learning can be viewed as a nonlinear function for representation learning with the deep hidden layers[23–25]. With this idea, autoencoder forms an unsupervised learning framework for learning the compressed representation in the hidden space of the input data[22,26,27]. An autoencoder consists of two modules: an encoder module and a decoder model. The encoder takes the input data and compresses it into a compressed representation in a lower-dimensional space, and the decoder then reconstructs the original inputs from the compressed representation. And the goal is to minimize the reconstruction error, as illustrated in Eq. (8).

$$f_\theta : \mathcal{X} \to \mathcal{H}$$
$$g_\theta : \mathcal{H} \to \mathcal{X} \quad (8)$$
$$\theta = \arg\min_\theta \|x - g(f(x))\|$$

Where $\mathcal{X}$ is the space of the inputs, $\mathcal{H}$ is the space of hidden states. In the autoencoder framework, $f$ learns to map the inputs to the compressed representation $f(x)$, and $g$ learns to reconstruct the inputs $g(f(x))$ from the compressed representation, and the goal is to minimize the error between the original inputs $x$ and the reconstructed $g(f(x))$. The model is then trained by minimizing the difference between the input and the output, also known as the reconstruction error. The popularity of autoencoder can be attributed to its ability to learn meaningful representations of complex data without requiring explicit supervision.

By this way, this compressed representation promises to capture the most important underlying structure of the input data and removes the redundant information. And the ability of autoencoder to learn meaningful representations and the underlying structure of complex data has made it a valuable model in many areas. In SHM, the learned compressed representation and the reconstructed errors are used as the damage indicator[3], given the facts that a well-trained autoencoder on the SHM big data can be viewed as a data-driven agent model of the bridge structural performance, and the reconstruction error actually reflects the changes of the underlying structures of the input data and the condition varies of the bridge structure compared to the initial condition.

### 2.4 The robust damage identification model based on LSTM

The proposed model integrates the above introduced modules and forms a LSTM-structured autoencoder model. Two stages are included: the first is the LSTM-structured encoder module that learns the underlying relationships and the hidden compressed representations and the LSTM-structured decoder module that learns to reconstruct the inputs from the hidden representation. And linear layers are added to the LSTM cell to reshape the size of the outputs of the LSTM-structured encoder and decoder.

Inputs are the monitored time series data of all investigated channels, while some of them are masked as dropout to simulate the missing data occasion. According to the above mentioned, this is equivalent to add unbiased noises on the training dataset, thus missing data cases are augmented in the training process, leading to an easier way to train the robust model that considers missing data. In this way, the spatiotemporal relationship among the time scale of all investigated channels are learned from the augmented dataset, and missing data is simply viewed as dropout. And the proposed model remain unchanged when the missing data occasion really occurs which makes the proposed method robust to missing data cases.

## 3. Case Study

### 3.1 Dataset

Cables are one of the most critical and vulnerable components in a cable-stayed bridge that suffer cyclic loads and harsh environments[28,29], and cable tension force is the most direct indicator. Opensource cable tension dataset of an in-service cable-stayed bridge released in 1st IPC-SHM (http://www.schm.org.cn/#/IPC-SHM,2020/dataDownload) is used for the case study. As shown in Fig. 3, the bridge is a double-cable-plane cable-stayed bridge with 168 cables (84 pairs). All cables are allocated with load cells for the monitoring of the dynamic cable tension. The monitoring data of 14 cables in 10 days are released, with the sampling frequency of 2Hz. Cables are numbered from left to right as: SJS08 to SJS14 for the upstream side of the bridge and SJX08 to SJX14 for the downstream side of the bridge. And the dates of the released dataset are a weeklong in 2006 from 2006-05-13 to 2006-05-19, and three separate days on 2007-12-14, 2009-05-05, and 2011-11-01. One of the released 14 cables are intended to be damaged while missing data occurs to 3 cables in the year 2011.

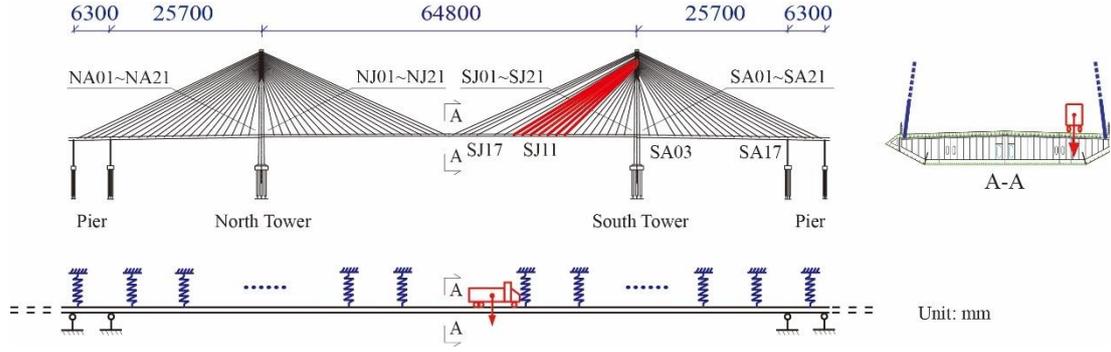

Figure 3. The investigated cable-stayed bridge
(Red cables denote the 14 cables in the released dataset)

A typical monitoring cable tension time series in one-day scale is illustrated in Fig. 4 (a) and details in 35 s scale is illustrated in Fig. 4 (b). Low frequency trend item induced by temperature and high frequency peak points item induced by vehicles can be observed. Further considering the dead load and noises, the monitored cable tension $T_{total}$ writes as: $T_{total} = T_d + T_e + T_v + T_r$, where $T_d$, $T_e$, $T_v$ and $T_r$ represent the effects of the dead load, the temperature, the vehicle and the noises, respectively. In the moving concentrated force assumption, the vehicle-induced cable tension can be expressed as:

$$\begin{bmatrix} T_{v,1} \\ T_{v,2} \\ \vdots \\ T_{v,M} \end{bmatrix} = \begin{bmatrix} d_{11} & d_{12} & \cdots & d_{1N} \\ d_{21} & d_{22} & \cdots & d_{2N} \\ \vdots & \vdots & \vdots & \vdots \\ d_{M1} & d_{M2} & \cdots & d_{MN} \end{bmatrix} \begin{bmatrix} F_1 \\ F_2 \\ \vdots \\ F_N \end{bmatrix} \qquad (9)$$

Where $T_{v,i}, T_{v,j}$ is the vehicle-induced cable tension item for $i$th and $j$th cable, $\boldsymbol{D}(\cdot) = [d_{mn}]_{M \times N}$ is the discretized flexibility matrix, and $F_n$ is the $n$th moving force loading on the discretized position.

Considering the cable tension of $i$th cable under a single vehicle,

$$T_{v,i}(t) = g_{il} \cdot F_n = \eta_i(x_n(t), y_n(t)) \cdot F_n \qquad (10)$$

Where $(x_n(t), y_n(t))$ represents the discretized location of the vehicle on the girder and $g_{il} = \eta_i(x_n(t), y_n(t))$ is the influence surface of the vehicle that is determined by the relative stiffness of stay cables and can be decoupled as the influence line on the longitudinal direction and the transverse direction $\eta_i(x_n(t), y_n(t)) = \eta_{xi}(x_n(t)) \cdot \eta_{yi}(y_n(t))$.[1] thus provides a single vehicle-based cable tension ratio as the damage indicator. Considering multiple vehicles on the bridge, which is more frequent in realistic, the vehicle-induced item of the $i$th and $j$th cable are:

$$T_{v,i}(t) = \sum_n F_n \cdot \eta_{xi}(x_n(t)) \cdot \eta_{yi}(y_n(t))$$
$$T_{v,j}(t) = \sum_n F_n \cdot \eta_{xj}(x_n(t)) \cdot \eta_{yj}(y_n(t)) \qquad (11)$$

## 3.2 Preprocessing

Vehicle-induced term of the cable tension is valuable since it provides a load-test style response information. Thus, it is necessary to decouple the multi-source effects and obtains the vehicle-induced term. However, due to the non-stationarity of this term as can be seen in Fig. 5(a,b), it's usually not easy to get the ideal term, smaller segmentations are suggested in this case[1,30]. Considering the sparsity of the vehicles on the bridge, a larger percentage of the observed data points are near the trend term that induced by temperature; thus violin plot is employed as the detrending technique in the preprocessing procedure.

A violin plot is a type of data visualization that combines the features of a box plot and a kernel density plot. It displays the distribution of a continuous variable across different levels of a categorical variable, as illustrated in Fig. 4(b). By visualizing the distribution of the data, it can be easily to identify that most observed datapoints are near the trend, thus the median value of a specified segmentation obtained by the violin plot is used as the trend term. Fig. 4 (b) shows the obtained trend term by this method in a 30-second segmentation, and Fig. 4(d) shows the obtained trend term of the whole-day scale by segmentations and interpolation, under the smooth assumption of the temperature-induced trend term.

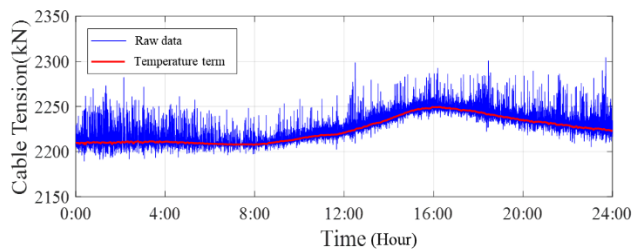
(a) Cable tension time series in one-day scale

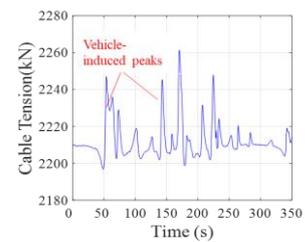
(b) Details in 350 s scale

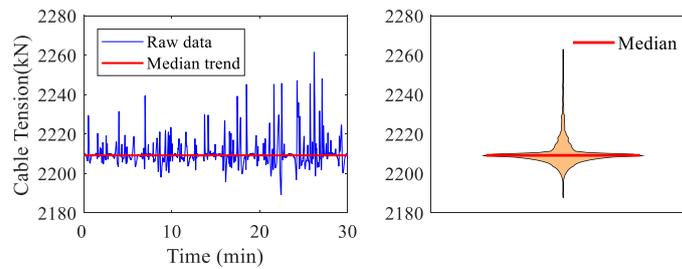
(c) Details in 30 minutes scale (**Left**: raw data and median trend, **Right**: violin plot)

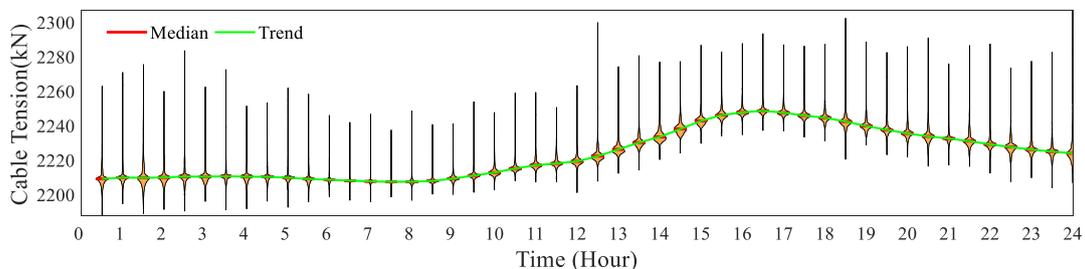
(d) Obtained temperature trend item

Figure 4. Multi-source of typical cable tension time series and trend item based on violin plots

### 3.3 Implementation details

For the dataset of the vehicle-induced cable tension $D = \{T_{v,i}(t)\}$, the vehicle loads are the same, therefore, the relations among them related to their influence surface and their relative stiffness only. Therefore, the model learned using the dataset of a specified period forms an agent model of that period, once the model changes in another period, the structural condition can be inferred as changed. And the reconstruction error predicted by the pretrained baseline model of a specific period (i.e., the early days of bridge operation) can be used as the damage indicator.

The proposed model learns the spatiotemporal relationships among the monitored data of all channels with the input and output nodes:

$$T_{v,i}(t) = AE_\theta \left( T_{v,j}(t+\tau) \right) \tag{12}$$

Following is the pseudo-code for the training procedure, the length of the LSTM inputs is the sequence of length $T$: $\{x^{(t_n+1)}, x^{(t_n+2)}, \cdots, x^{(t_n+T)}\}$, inputs $x^{(t_n)}: \{T_{v,i}^b(t_n)\} \in R^{\text{batch\_size} \times M}$ represents the batch_size of the vehicle-induced cable tension of M cables at $t_n$ step, outputs $\hat{y}^{(t)}$ represents the corresponding predictions, in the unsupervised learning way, target outputs are equal to inputs $y^{(t)} = x^{(t)}$. Adam optimization algorithm is employed to training the model.

---

Pseudo-code: LSTM-structured autoencoder training

---
1. Initialize parameters $\theta$;
2. Specifies batch_size, the length of the input $T$, learning rate $\alpha$;
3. For $k=1,\cdots, max\_iteration$
4. Randomly generates batch_size integers from $[1, N-T]$;
5. Generates normalized training set $x = \{T_v^{b,i}(t_n : t_n + T)\} \in R^{\text{batch\_size} \times T \times M}$, $target\_y = x$, randomly sets dimensions $M_1 \in [1, M]$ in $x$ to 0 to obtain $train\_x = \bar{x}$;
6. Updating model parameters using AdamOptimizer.

---

In this task, units for the inputs and outputs are 14, corresponding the the 14 channels of the monitoring dataset. The LSTM-structured encoder and the decoder cell are both with 3 layers and 32 units in each layer. Linear layers of $layer$ 1: $\mathbb{R}^{32} \to \mathbb{R}^5$ and $layer$ 2: $\mathbb{R}^{32} \to \mathbb{R}^{14}$ are added to the encoder and the decoder respectively to satisfies the shapes of the hidden layer and the outlayer. Other hyperparameters are set as: $T = 2400$, $max\_iteration=100{,}000$, $batch\_size = 30$ and $\alpha=0.005$, and MSE error are employed. The whole model is build on TensorFlow and optimizer is chosen as Adam Optimizer. An autoencoder structured by a 3-layer DNN with [64, 32, 5] hidden units and a decoder structured by a 3-layer DNN with [5, 32, 14] hidden units are also developed as a comparison model.

## 4. Results and Discussions

0-12 cables are simulated to be dropped (corresponding to 0-85.7% missing rate) in the training procedure. Figure 5 (a) shows the loss curves of DNN-structured and the LSTM-structured autoencoder model trained with a 50% data loss rate (i.e., 7 randomly missing cable data are dropped out). Figure 5 (b) shows the training errors of DNN and LSTM networks under different data loss rates. It can be observed from Figure 6 that in the case of data loss, the training performance of DNN and LSTM networks increases exponentially with the increase of the data loss rate (i.e., the number of missing cables k). Compared with the DNN-structured AE model, the LSTM-structured AE model can form a memory of historical data within the network, thereby extracting spatiotemporal correlation features of cable forces in the cable group. Therefore, under the missing date occasion, the performance of the LSTM-structured AE model and spatiotemporal correlation of cable tension far exceeds that of the DNN-structured AE model and spatial correlation only.

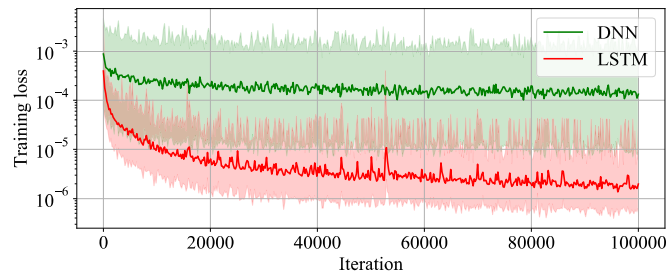

(a) Training losses of DNN and LSTM-structured AE model under 50% missing rate

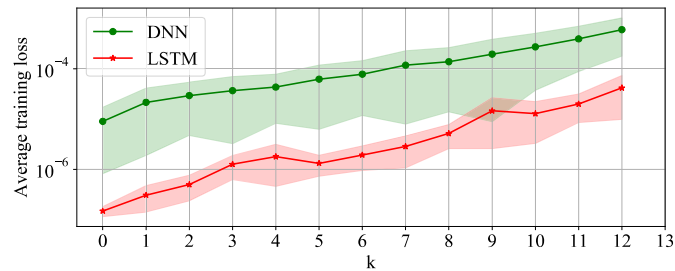

(b) Training loss of DNN and LSTM-structured AE model under varing missing rate

Figure 5. Comparison of DNN-structured and LSTM -structured AE model

Fig. 6 shows the prediction results of the pre-trained LSTM network for cable tension on November 1, 2011. While cable tension of SJX08, SJS13, and SJX13 are real missing, cable tension of SJS08, SJS09, SJS11, SJX11, SJX12, SJS13, and SJX13 are dropped out. Eight cable forces were actually missing in the input data. With the pretrained model, the prediction is illustrated in Fig 6. Indicating that the LSTM-structured AE model can reconstruct the missing data and diagnose cable health based on the pretrained agent model of the cable group's initial performance. The real cable tension value of cable SJS11 in Fig. 6(c) is lower than the benchmark prediction value, indicating a decrease in the actual carrying capacity of the cable and thus indicating the cable damage, which is consistent with the cable state evaluation results released in [14].

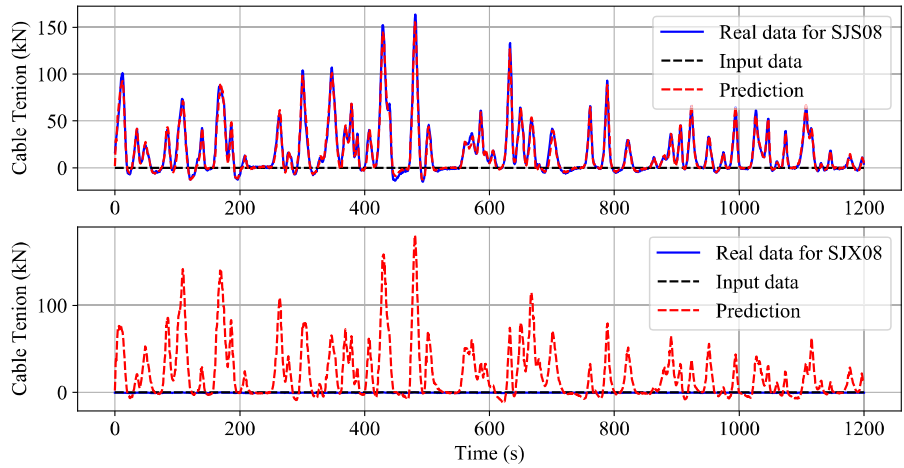

(a) SJS08 vs SJX08 (where SJX08 is loss and SJS08 is assumed to be loss)

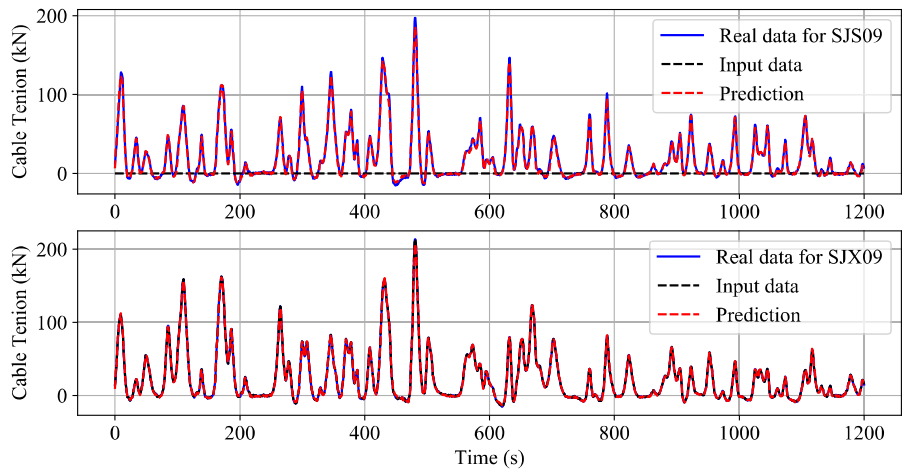

(b) SJS09 vs SJX09 (where SJS09 is assumed to be loss)

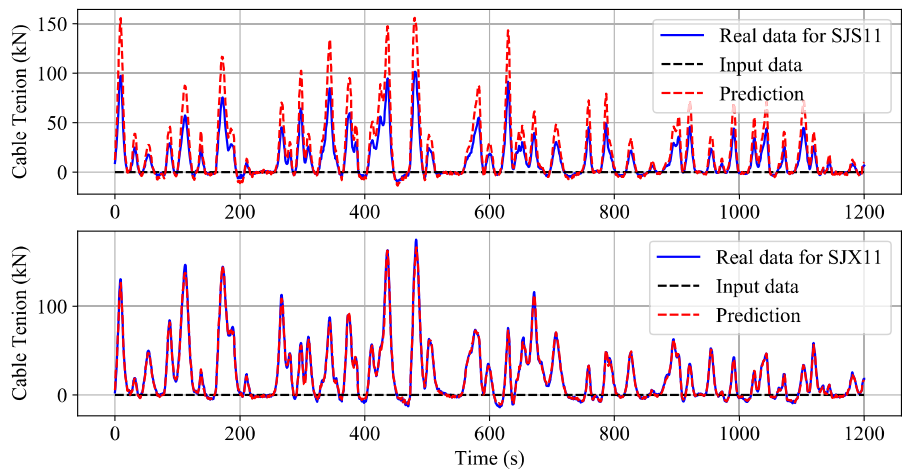

(c) SJS11 vs SJX11 (SJS11 and SJX11 are assumed to be loss, and SJS11 is damage)

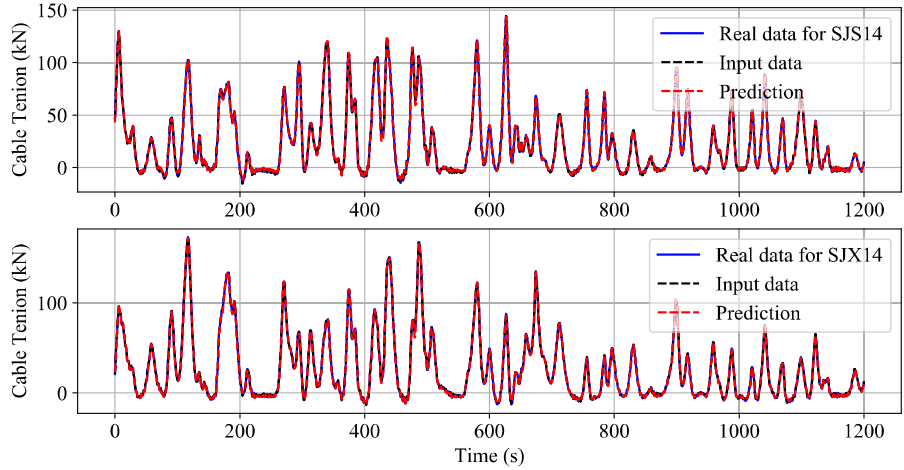

(d) SJS14 vs SJX14

Figure 6. Prediction results of DNN and LSTM in 50% data loss rate case (2011-11-01)

Fig. 7 shows the diagnosis results of the cable group based on the 3-σ criterion. The results indicate that only cable SJS11 is damaged based on the standard values of the prediction error calculated using the pre-trained LSTM-structured AE model, while the other cables are healthy condition, which is consistent with the cable state evaluation results in [14].

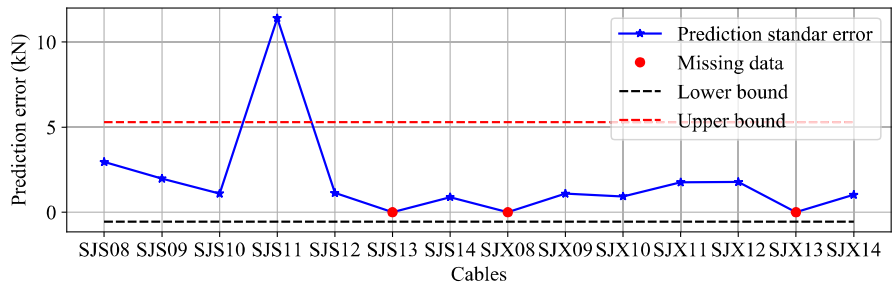

Figure 7. Damage identification based on the reconstruction error

## 5. Conclusion

This paper discussed the missing data issue in SHM and the dropout mechanism for the consideration of missing data in data-driven approaches of SHM. Results show that we can view the missing data with dropped input units to develop the robust model for damage identification and other tasks, instead of the standard flow of missing data imputation pre-processing and downstream tasks. Following conclusions can be also obtained.

Unsupervised learning representations learned based on LSTM-structured AE model forms a baseline agent model for the correlationship of a group of cables, and the reconstruction error can be used as the damage indicator. The baseline model for the cable group establishes a many-to-many spatiotemporal mapping model for cables in the group. Which improves the efficiency of structural health monitoring data processing and health diagnosis.

The unsupervised learning representation of LSTM-structured AE model and the dropout mechanism employed in this paper can loosen the requirements of data quality (missing data occasion) while achieving an accurate and robust baseline agent model. And with this model, missing data imputation and damage identification can be conducted in the unified way at the meantime.

# Acknowledgements

This work is financially supported by the National Natural Science Foundation of China (Grant Nos. (Grant Nos. 52208311, 51921006 and 52192661). The authors would like to thank the organizations of the International Project Competition for SHM (IPC-SHM 2020) ANCRiSST, Harbin Institute of Technology (China), and University of Illinois at Urbana-Champaign (USA) for their generously providing the invaluable data from actual structures.